\documentclass{article}

\usepackage{arxiv}
\usepackage[utf8]{inputenc} 
\usepackage{csquotes}
\usepackage[T1]{fontenc}    
\usepackage{hyperref}       
\usepackage{url}            
\usepackage{booktabs}       
\usepackage{amsfonts}       
\usepackage{nicefrac}       
\usepackage{microtype}      
\usepackage{lipsum}
\usepackage{graphicx}
\usepackage{subfigure}
\usepackage{comment}
\usepackage[english]{babel} 
\usepackage{microtype} 
\usepackage{amsmath,amsfonts,amsthm} 
\usepackage[svgnames]{xcolor} 
\usepackage[hang, small, labelfont=bf, up, textfont=it]{caption} 
\usepackage{booktabs} 
\usepackage{lastpage}
\usepackage{graphicx} 
\usepackage{enumitem} 
\setlist{noitemsep} 
\usepackage[capitalise]{cleveref}
\usepackage[style=numeric, sorting=none]{biblatex}
\usepackage{comment}
\usepackage{sectsty} 
\allsectionsfont{\usefont{OT1}{phv}{b}{n}} 
\usepackage[normalem]{ulem}
\usepackage{venndiagram}
\usepackage{siunitx}

\usepackage{amsmath,amssymb,amsfonts}
\usepackage{algorithmic}
\usepackage{graphicx}
\usepackage{textcomp}
\usepackage[ruled]{algorithm2e}
\usepackage{bm}
\usepackage{xcolor}
\usepackage{array}
\usepackage{comment}

\SetKwInput{KwInput}{Input}                
\SetKwInput{KwOutput}{Output}              
\SetKwInput{kwConstraint}{Constraint}      
\SetKwFunction{EDPLS}{edpls}
\SetKwProg{Fn}{Function}{:}{End}

\newcommand{\bd}[1]{ \mbox{\boldmath $\mathrm{#1}$} }
\newcommand{\scr }{ \scriptsize }

\newcommand{\tr}{ \scr \mbox{T} } 

\newcommand{\dblvert}{ \left| \left|  }
\newcommand{\dbrvert}{ \right| \right| }
\newcommand{\fronorm}[1]{ \dblvert #1 \dbrvert_{\scr F}}

\newcommand{\twonorm}[1]{ \dblvert #1 \dbrvert_2 }

\newcommand{\epsdel}{^{(\epsilon, \delta)}}

\title{$(\epsilon, \delta)$-differentially private partial least squares regression}

\author{
  Ramin Nikzad-Langerodi\footnote{Corresponding author}\\
  Software Competence Center Hagenberg (SCCH) GmbH\\
  Hagenberg, Austria\\ 
  \texttt{ramin.nikzad-langerodi@scch.at}\\
  \And
  Mohit Kumar\\
  Faculty of Computer Science and Electrical Engineering \\
       University of Rostock, Germany\\
        and\\
  Software Competence Center Hagenberg (SCCH) GmbH\\
  Hagenberg, Austria\\ 
  \texttt{mohit.kumar@uni-rostock.de}
  \And
  Du Nguyen Duy\\
  Software Competence Center Hagenberg (SCCH) GmbH\\
  Hagenberg, Austria\\ 
  \texttt{du.nguyen.duy@scch.at}
  \And
  Mahtab Alghasi\\
  Software Competence Center Hagenberg (SCCH) GmbH\\
  Hagenberg, Austria\\ 
  \texttt{mahtab.alghasi@scch.at}}

\begin{document}

\maketitle

\begin{abstract}
As data-privacy requirements are becoming increasingly stringent and statistical models based on sensitive data are being deployed and used more routinely, protecting data-privacy becomes pivotal. Partial Least Squares (PLS) regression is the premier tool for building such models in analytical chemistry, yet it does not inherently provide privacy guarantees, leaving sensitive (training) data vulnerable to privacy attacks. To address this gap, we propose an $(\epsilon, \delta)$-differentially private PLS (edPLS) algorithm, which integrates well-studied and theoretically motivated Gaussian noise-adding mechanisms into the PLS algorithm to ensure the privacy of the data underlying the model. Our approach involves adding carefully calibrated Gaussian noise to the outputs of four key functions in the PLS algorithm: the weights, scores, $X$-loadings, and $Y$-loadings. The noise variance is determined based on the global sensitivity of each function, ensuring that the privacy loss is controlled according to the $(\epsilon, \delta)$-differential privacy framework. Specifically, we derive the sensitivity bounds for each function and use these bounds to calibrate the noise added to the model components. Experimental results demonstrate that edPLS effectively renders privacy attacks, aimed at recovering unique sources of variability in the training data, ineffective. Application of edPLS to the NIR corn benchmark dataset shows that the root mean squared error of prediction (RMSEP) remains competitive even at strong privacy levels (i.e., $\epsilon=1$), given proper pre-processing of the corresponding spectra. These findings highlight the practical utility of edPLS in creating privacy-preserving multivariate calibrations and for the analysis of their privacy-utility trade-offs.
\end{abstract}


\section{Introduction}\label{introduction}
Data-privacy has become an increasingly important societal topic as organizations strive to harness the power of machine learning (ML) and AI on aggregated data. Over the past two decades, significant strides have been made in establishing methods that allow the derivation of valuable insights from data while safeguarding data-privacy. Techniques like differential privacy (DP) and federated learning (FL) represent key advancements in this area, embodying the delicate balance between maximizing data utility and ensuring security. Following the seminal work of Dwork \emph{et al.} \cite{dwork2006calibrating}, who introduced differential privacy as a theoretical framework for developing methods with privacy guarantees, i.e., where an algorithm’s output does not depend on individual samples in the underlying dataset, the field has evolved to include diverse applications and methodologies pertaining to machine- and deep learning \cite{liu2021machine}. 

Privacy-enhancing techniques (PETs) have been (primarily) explored in the realm of unsupervised/exploratory data analysis and multivariate statistics. The Sub-Linear Queries (SuLQ) framework by Blum \emph{et al.} proposes the addition of $i.i.d$ Gaussian noise to the sample covariance matrix to derive privacy-preserving principal components (PCs) \cite{blum2005practical}. Chaudhuri \emph{et al.} noted that, while obeying DP, the corresponding PCs might contain complex entries since the symmetry of the covariance matrix is broken by adding the (non-symmetric) noise matrix, and proposed a modification involving a symmetric noise adding mechanism \cite{chaudhuri2012near}. Despite the fact that a relatively small amount of noise is often sufficient for privacy-preservation, PETs usually come with privacy-accuracy trade-offs~\cite{Kumar/IWCFS2019,KUMAR202187,Geng2018OptimalNM,balle18a,Gupte:2010:UOP:1807085.1807105,7345591,7093132,7353177,kumar2023differentially,10.1613/jair.1.15071} - especially for small datasets. Chai \emph{et al.} proposed a simple approach that claims lossless privacy-protection in PCA by means of removable random masks \cite{10.1145/3534678.3539402}. In brief, data is locally masked by multiplication with some full-rank random matrix prior to concatenation with similarly masked datasets from other parties. The private scores (in horizontal FL) or loadings (in vertical FL) are locally "decrypted" following the decomposition of the concatenated dataset (using the inverse of the corresponding random masks). This approach has been explored by Nguyen Duy \emph{et al.} in the scope of vertical federated process modeling by means of unsupervised, PCA-based \cite{nguyen2023fedmspc} and supervised, Partial least squares (PLS) based \cite{NGUYENDUY2024103229} multivariate statistical process control (MSPC). However, although privacy attacks have not yet been reported, strictly speaking, the removable random masks approach employed in these federated singular values decomposition (SVD) based techniques does not obey DP.    

In the present study, we explore how private information might be leaked in multivariate calibration on horizontally concatenated data originating from different data holders. Rather than focusing (exclusively) on DP, whereas a model is usually deployed on a server and provides "noisy" outputs to prevent an adversary from gaining information about individual data points in the training set, we investigate how the data holders might recover information in terms of (unique) sources of variability present in the other parties' training data based upon common PLS model components. We then propose a Gaussian noise adding mechanism according to \cite{dwork2014algorithmic} and \cite{dwork2014analyze} for the PLS1 algorithm, to ensure $(\epsilon, \delta)$-differential privacy of the corresponding model. Finally, we apply the proposed $(\epsilon, \delta)$-differentially private PLS (i.e., edPLS) model to simulated and real-world data in the context of spectroscopic data analysis and multivariate calibration.

Our experiments on simulated data show how edPLS renders privacy attacks, aimed at recovering unique sources of variability underlying the training data, ineffective. In addition, we found that achieving strong privacy guarantees while maintaining the utility of a multivariate calibration on the near-infrared (NIR) corn benchmark requires adequate (privacy-enabling) pre-processing of the corresponding spectra.

\section{Theory}
\label{theory}

\subsection{Notation}
\label{sec:notation}

All matrices are presented in bold upper-case letters, and one-dimensional arrays are presented in bold lower-case letters. The superscript $^T$ denotes the transpose of matrices and one-dimensional arrays and $^{-1}$ the unique inverse. We consider the standard PLS1 model based upon the multivariate matrix $\bd X\in\mathcal{X}\subseteq\mathbb{R}^{n \times m}$ of predictors and univariate vector $\bd y\in\mathcal{Y}\subseteq\mathbb{R}^{n\times 1}$ holding the corresponding responses, i.e.,  $f(\bd x) = \bd x^T\bd W(\bd P^T\bd W)^{-1}\bd{q}$, where $\bd W, \bd P$ and $\bd q$ hold the weights, $X$- 
and $Y$-loadings, respectively, and the regression coefficients
\begin{equation}
\label{eq:reg_coeff}
\bd b = \bd W(\bd P^T\bd W)^{-1}\bd{q}.
\end{equation}
We further use $\Pr[\cdot]$, $\twonorm{\cdot}$ and $\mathcal{N}(0, \sigma^2)$ to denote probability, the $\ell_2$-norm, and Normal distribution with 0 mean and variance of $\sigma^2$, respectively. By \emph{neighboring datasets} we mean two datasets $D=(\bd X, \bd y)$ and $D'=(\bd X', \bd y')$, where one sample (row) has been removed from the former to derive the latter, i.e., $\bd X\in \mathbb{R}^{n \times m}, \bd X'\in \mathbb{R}^{(n-1)\times m}$ and $\bd y\in\mathbb{R}^{n \times 1}, \bd y'\in\mathbb{R}^{(n-1) \times 1}$ \cite{zhao2019reviewing}. $\bd x_i$ denotes the the $i$-th row of $\bd X$.  We use $\sup_{D,D'}\|\cdot\|_2$ to denote the supremum of the $\ell_2$-norm over neighboring datasets, i.e., the worst-case change of the norm, if a single $(\bd x, y)$-pair is removed (or added). We assume that $\bd X$ and $\bd y$ are mean-centered.

\subsection{Differential Privacy}
\label{sec: differential_privacy}
Differential privacy is a rigorous mathematical framework designed to provide strong privacy guarantees of an algorithm $\mathcal{A}$ processing statistical data $D$. The concept is characterized by two parameters, $\epsilon$ and $\delta$, which together quantify the privacy loss of $\mathcal{A}$. An algorithm is said to be $(\epsilon, \delta)$-differentially private if, for any two datasets $D$ and $D'$ that differ by a single element (i.e., one row/sample), the probability that the algorithm produces a given output $S$ does not change by more than a multiplicative factor of $e^\epsilon$ and an additive factor of $\delta$, i.e., 

\begin{equation}
\label{eq:diff_privacy}
\Pr[\mathcal{A}(D) \in S] \leq e^\epsilon \Pr[\mathcal{A}(D') \in S] + \delta.
\end{equation}

In simpler terms, $\epsilon$ controls the privacy-accuracy trade-off, with smaller values indicating stronger privacy at the cost of reduced accuracy. The parameter $\delta$ represents the probability of the privacy guarantee being violated, with smaller values indicating a lower failure probability. If $\epsilon = 0$ (and $\delta$ small), $\mathcal{A}$ returns the same output on $D$ and $D'$ with high probability and is thus largely independent of individual database entries, hence protecting their privacy.

As outlined in Dwork \emph{et al.} \cite{dwork2014analyze}, given some (vector-valued) function $f:\mathcal{D}\rightarrow \mathbb{R}^m$ operating on databases, addition of independently drawn random noise distributed as $\mathcal{N}(0,\sigma^2)$, with 
\begin{equation}
\label{eq:noise_std}
\sigma = \Delta f\frac{\sqrt{2\ln(1.25/\delta)}}{\epsilon}
\end{equation}
to each output $f(D)$ ensures $(\epsilon, \delta)$-differential privacy\footnote{As shown in \cite{balle18a}, the corresponding Gaussian mechanism is differentially private only for $\epsilon\leq1$ and $\delta\in[0, 1]$.}. $\Delta f$ in Eq. \eqref{eq:noise_std} is called the \emph{sensitivity} and represents a "worst case" measure for how the output of $f$ changes when a single entry in $D$ is changed, i.e., 
\begin{equation}
\label{eq:sensitivity}
\Delta f = \sup_{D, D'}||f(D)-f(D')||_2.
\end{equation}
As will be discussed in section \ref{sec:edpls}, the main task of ensuring that an algorithm is $(\epsilon, \delta)$-differential private is to carefully "calibrate" the noise variance according to the sensitivity of the corresponding function.

\subsection{Problem Statement}
\label{sec:problem}
\begin{figure}[ht!]
    \centering
    \includegraphics[width=0.9\linewidth]{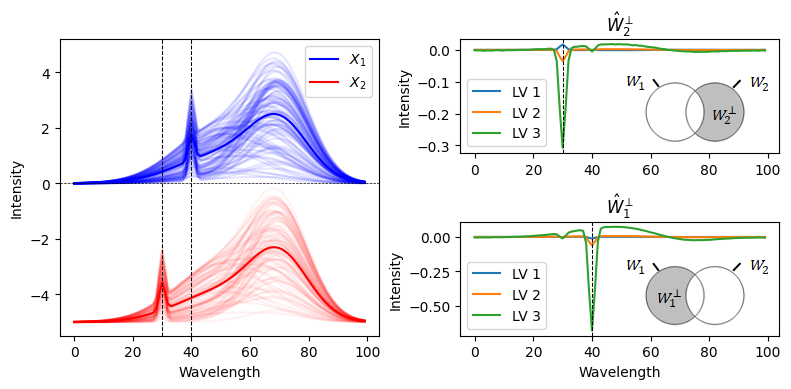}
    \caption{Motivating example. If the owners of two datasets $(\bd X_1, \bd y_1)$ and $(\bd X_2, \bd y_2)$ (left) perform a PLS regression on the concatenated dataset, recovery of the unique sources of variability in the other's $X$-data is trivial (right). This is true even if both parties don't have access to each other's raw data and concatenation/modeling takes place on a remote server.}
    \label{fig:motivating_example}
\end{figure}

For illustrative purposes, we consider a scenario with two data holders $\mathcal{D}_1$ and $\mathcal{D}_2$ that own datasets $(\bd X_1, \bd y_1)$ and $(\bd X_2, \bd y_2)$, which they share with a (trusted) computation service provider (CSP). The CSP then horizontally concatenates the data such that $\bd X = [\bd X_1; \bd X_2]$ and $\bd y = [\bd y_1; \bd y_2]$, performs a PLS regression of $\bd y$ on $\bd X$ and broadcasts the PLS model components (see Eq.\ref{eq:reg_coeff}) back to both data holders.
Given this simplified learning setting, identification of unique sources of variability underlying the $X$-data provided by the other data holder is straightforward (Figure \ref{fig:motivating_example}). To this end, the $i$-th data holder fits a local PLS model using his data ($\bd X_i, \bd y_i$) with the same number of LVs as the global model (obtained by the CSP) and deflates the global weights using the local weight matrix by means of a Gram-Schmidt process, i.e.,  
\begin{align} 
\begin{aligned}
\hat{\bd W}_j^{\perp} &= \bd W - \text{proj}_{\bd W_i}(\bd W) \\ &= \bd W - \bd W_i(\bd W_i^T\bd W_i)^{-1}\bd W_i^T\bd W & (i\neq j),
\end{aligned}
\label{eq:w_perp}
\end{align}
where $\bd W$ and $\bd W_i$ denote the global and local weights, respectively. $\hat{\bd W}_j^{\perp}$ provides an estimate of the unique sources of variability orthogonal to $\bd W_i$, yet in the space spanned by the columns of $\bd W$\footnote{This is because the columns of $\bd W$ and $\bd W_i$ are both in the row-space of $\bd X$.}. An example using a simulated data set is provided in Figure \ref{fig:motivating_example}, whereas LV 3 of $\hat{\bd W}_j^{\perp}$ clearly reveals the (unique) peak present in the other data holder's spectra. The same approach can be used to reveal unique components in the loadings (not shown). In order to protect the model from such privacy attacks, ensuring that the data underlying the model remains private, i.e., $(\epsilon, \delta)$-differential privacy, we propose a Gaussian mechanism for PLS regression in the next section. 

\subsection{$(\epsilon, \delta)$-Differentially private PLS}
\label{sec:edpls}
As outlined in section \ref{sec: differential_privacy}, ensuring $(\epsilon, \delta)$-differential privacy amounts to calibrating the variance of the noise added to a function's output, according to its sensitivity, prior to releasing the output. The PLS algorithm has four distinct functions to consider (see Algorithm 1):

\begin{align}
\begin{aligned}
\bd w:\mathcal{X}&\times \mathcal{Y}\rightarrow\mathbb{R}^m, \quad & \bd w(\bd X, \bd y) &= \bd X^T\bd y\\
\bd t:\mathcal{X}&\times \mathcal{W}\rightarrow\mathbb{R}^n, \quad &\bd t(\bd X, \bd w) &= \bd X \bd w/\twonorm{\bd w}\\
\bd p:\mathcal{X}&\times \mathcal{T}\rightarrow\mathbb{R}^m, \quad &\bd p(\bd X, \bd t) &= \bd X^T\tilde{\bd t} /(\tilde{\bd t}^T\tilde{\bd t})\\
c:\mathcal{Y}&\times \mathcal{T}\rightarrow\mathbb{R}, \quad &c(\bd y, \bd t) &= \bd y^T\tilde{\bd t} /(\tilde{\bd t}^T\tilde{\bd t}), 
\end{aligned}
\end{align}

with $\mathcal{X} \subseteq \mathbb{R}^{n\times m}$, $\mathcal{Y} \subseteq \mathbb{R}^{n}$, $\mathcal{W} \subseteq \mathbb{R}^{m}$, $\mathcal{T} \subseteq \mathbb{R}^{n}$ and $\mathcal{C} \subseteq \mathbb{R}$.
The vector-valued function $w$ takes inputs $\bd X$ and $\bd y$, and returns the covariance $\bd w = \bd X^T\bd y$. Its sensitivity can be upper bounded as follows (see Appendix for derivation):

\begin{align}
\label{eq:w_sensitivity}
\begin{aligned}
\Delta w &= \sup_{D, D'} \|\bd w(\bd X, \bd y) - \bd w(\bd X', \bd y')\|_2 \\
         &= \sup_{(\mathbf x, y)} |y|\|\bd x\|_2
\end{aligned}
\end{align}

We estimate the corresponding terms (as conservatively as possible) such that

\begin{align}
\label{eq:supremum_sample}
\begin{aligned}
\sup_{\mathbf x}\|\bd x\|_2 := \max_i \|\bd x_i\|_2 \quad \text{and} \quad
\sup_{y} |y| := y_{\text{max}},
\end{aligned}
\end{align}

where $y_{\text{max}}$ denotes the maximum value in $\bd y$ and $\max_i \|\bd x_i\|_2$ the maximum row-norm in $\bd X$ (after mean centering). We resort to the approach proposed in \cite{balle18a} to compute the noise variance $\sigma_w^2$ instead of using Eq. \eqref{eq:noise_std}, to guarantee $(\epsilon, \delta)$-differential privacy in both low ($\epsilon > 1$) and high ($\epsilon\leq 1$) privacy regimes. The global sensitivity $\Delta w$ is estimated using Eqs. \eqref{eq:supremum_sample}. Finally, the $(\epsilon, \delta)$-differentially private weight vector $\bd w\epsdel = \bd w + \bd v_w$ is released following addition of the vector $\bd v_w$, holding Gaussian noise sampled from $\mathcal{N}(0, \sigma_w^2)$. 

The Gaussian mechanism for the scores is calibrated using the upper bound on the sensitivity of the corresponding function $\bd t$ (see Appendix), i.e.,

\begin{align}
\label{eq:t_sensitivity}
    \begin{aligned}
    \Delta t &\leq \sup_{\mathbf x} \|\bd x\|_2. 
    \end{aligned}
\end{align}

The noise variance $\sigma_t^2$ is then computed as described above and the $(\epsilon, \delta)$-differentially private scores vector $\bd t\epsdel = \bd t + \bd v_t$ released following addition of the Gaussian noise vector $\bd v_t$ obtained by sampling $\mathcal{N}(0, \sigma_t^2)$. Note that we normalize the scores to unit length prior to the computation of loadings, i.e., $\tilde{\bd t} = \bd t/\|\bd{t}\|_2$. The sensitivity of the loadings functions $\bd p$ and $c$ are  bounded as follows (see Appendix):

\begin{align}
\label{eq:p_sensitivity}
    \begin{aligned}
    \Delta p & \leq \sup_{\mathbf x} \|\bd x\|_2. 
    \end{aligned}
\end{align}
and
\begin{align}
\label{eq:c_sensitivity}
    \begin{aligned}
    \Delta c \leq \sup_{y} |y|. 
    \end{aligned}
\end{align}
Again, the sample bounds in Eq.\eqref{eq:supremum_sample} and the algorithm from \cite{balle18a} are used to compute the noise variances, followed by sampling the vector 
$\bd v_p\sim\mathcal{N}(0, \sigma_p^2)$ and scalar $v_c\sim\mathcal{N}(0, \sigma_c^2)$ to be added for differential private release of $\bd p\epsdel=\bd p + \bd v_p$ and $c\epsdel = c + v_c$, respectively. 

To conclude one iteration of the $(\epsilon, \delta)$-differentially private PLS algorithm, $\bd w\epsdel$, $\bd t\epsdel$, $\bd p\epsdel$ and $c\epsdel$ are stored and the matrix $\bd X$ and vector $\bd y$ deflated using the non-private scores $\bd t$ and loadings $\bd p$ and $c$. After $k$ components have been fitted, the regression coefficients are calculated according to Eq.\eqref{eq:reg_coeff} using the matrices $\bd W\epsdel$, $\bd P\epsdel$ and vector $\bd c\epsdel$ holding the privately released model components. Note that either noisy or non-private versions of weights, scores, and loadings are used during the computations and that weights and scores are normalized before and after noise addition (see Algorithm 1). 

\begin{algorithm*}
\DontPrintSemicolon
  
  \KwInput{$\bd{X}$, $\bd{Y}$, $k$, $\epsilon$, $\delta$}
  \KwOutput{$\bd{T}\epsdel$, $\bd{c}\epsdel$, $\bd{W}\epsdel$, $\bd{P}\epsdel$, $\bd{b}\epsdel$}
  \Fn{\EDPLS{$\bd{X}$, $\bd{Y}, k$, $\epsilon$, $\delta$}}{
            
        Initialize matrix $\bd{E} := \bd{X}$, and vector $\bd{f} := \bd{y}$\\
        
        \For{$j = 1 \rightarrow k$}{
            Calculate PLS weight vector $\bd w$: 
            \begin{equation*}
            \bd{w} := \frac{\bd{E}^T \bd{f}}{\|\bd{E}^T \bd{f}\|_2}
            \end{equation*}\\
            Compute sample bounds in Eq. \eqref{eq:supremum_sample} using $\bd E$ and $\bd f$:
            \begin{align*}
            y_{\text{max}} := \max_i |\bd f_i|, \quad \max_i \|\bd x_i\|_2 := \max_i \|\bd {E}_i\|_2
            \end{align*}\\
            Calculate $\Delta w$ according to Eq. \eqref{eq:w_sensitivity}\\
            Calculate noise variance $\sigma_w^2$ using Algorithm 1 in \cite{balle18a}.\\
            Sample the $m \times 1$ vector $\bd{v}_w$ from the distribution $\mathcal{N}(0, \sigma_w^2)$ and add it to \bd{w}:
              \begin{equation*}
                \bd{w}\epsdel := \bd{w} + \bd{v}_w
              \end{equation*}\\
            Normalize $\bd w\epsdel$ to unit length:
            \begin{equation*}
                \bd w\epsdel := \frac{\bd w\epsdel}{\twonorm{\bd w\epsdel}}
            \end{equation*}\\
            Calculate (non-private) unit length-scaled scores $\bd{t}$:
            \begin{align*}
                \bd t := \frac{\bd{E}\bd{w}}{\|\bd E\bd w\|_2}
            \end{align*}\\
            Calculate $\Delta t$ according to Eq. \eqref{eq:t_sensitivity} using the sample upper bounds in Eq. \eqref{eq:supremum_sample}.\\
            Calculate noise variance $\sigma_t^2$ according to \cite{balle18a}.\\
            Sample the $n \times 1$ vector $\bd{v}_t$ from the distribution $\mathcal{N}(0, \sigma_t^2)$ and add it to $\bd{t}$:
            \begin{equation*}
                \bd{t}\epsdel := \bd{t} + \bd{v}_t
            \end{equation*}\\
            Normalize scores:
            \begin{align*}
            \bd t\epsdel := \frac{\bd t\epsdel}{\|\bd t\epsdel\|_2}
            \end{align*}\\
            Calculate $X$- and $Y$-loadings $\bd{p}$ and $c$:
            \begin{align*}
                \bd{p} := \frac{\bd E^T\bd t}{\bd t^T\bd t}, \quad c := \frac{\bd{f}^T\bd{t}}{\bd t^T\bd t}
            \end{align*}\\
            Calculate $\Delta p$ and $\Delta c$ according to Eq. \eqref{eq:p_sensitivity} and Eq. \eqref{eq:c_sensitivity} using the sample upper bounds in Eq. \eqref{eq:supremum_sample}.\\
            Calculate noise variances $\sigma_p^2$ and $\sigma_c^2$ according to \cite{balle18a}.\\
            Sample the vector $\bd{v}_p$ from the distribution $\mathcal{N}(0, \sigma_p^2)$ and scalar $v_c$ from $\mathcal{N}(0, \sigma_c^2)$.\\
            Add noise to $\bd{p}$ and $c$:
            \begin{equation*}
                \bd{p}\epsdel := \bd{p} + \bd{v}_p, \quad c\epsdel := c + v_c
            \end{equation*}\\
            Deflate the data matrices $\bd{E}$ and $\bd{f}$ using non-private scores and loadings:
            \begin{equation*}
                \bd{E} := \bd{E} - \bd{t}\bd{p}^T, \quad \bd{f} := \bd{f} - \bd{t} c
            \end{equation*}\\
            Add $\bd{w}\epsdel$, $\bd{t}\epsdel$, $\bd{p}\epsdel$ as columns to $\bd{W}\epsdel$, $\bd{T}\epsdel$, $\bd{P}\epsdel$ and the scalars $c\epsdel$ to vector $\bd{c}\epsdel$, respectively.\\
        }
        
        Compute regression coefficients $\bd b\epsdel$:
            \begin{equation*}
                \bd b\epsdel := \bd{W}\epsdel(\bd{P}{\epsdel}^T\bd{W}\epsdel)^{-1}\bd{c}\epsdel
            \end{equation*}
  }
\caption{$(\epsilon,\delta)$-Differentially Private PLS}
\label{alg:ed_pls}
\end{algorithm*}

\section{Experimental}\label{experiments}

\subsection{Datasets}\label{dataset-descr}

\paragraph{Simulated Data Set:} We generated four Gaussian signals with $m=100$ points to represent the analyte and three interferents using different parameters for the signal center $\mu$, width $\sigma$ and height $h$ such that for

\begin{itemize}
    \item \textbf{Analyte Signal ($\bd s_1$)}: $\mu = 50$, $\sigma = 15$, $h=8$.
    \item \textbf{Interferent 1 Signal ($\bd s_2$)}: $\mu = 70$, $\sigma = 10$, $h= 10$.
    \item \textbf{Interferent 2 Signal ($\bd s_3$)}: $\mu = 40$, $\sigma = 1$, $h=0.5$.
    \item \textbf{Interferent 3 Signal ($\bd s_4$)}: $\mu = 30$, $\sigma = 1$, $h=0.5$.
\end{itemize}
$n=100$ random concentrations $\bd c_1, \bd c_2, \bd c_3, \bd c_4$ were generated from a uniform distribution in the half-open interval $[0, 10)$ for each of the signals. Then two spectra-like datasets were generated such that 
\begin{align*}
\begin{aligned}
\bd X_1 & = \bd C_1\bd S_1^T \;\; , \;\; \bd y_1 = \bd c_1\\
\bd X_2 & = \bd C_2\bd S_2^T \;\; , \;\; \bd y_2 = \bd c_1,
\end{aligned}
\end{align*}
with $\bd C_1 = [\bd c_1, \bd c_2, \bd c_3]$, $\bd C_2 = [\bd c_1, \bd c_2, \bd c_4]$,  $\bd S_1 = [\bd s_1, \bd s_2, \bd s_3]$ and $\bd S_2 = [\bd s_1, \bd s_2, \bd s_4]$. The individual concentration vectors $\bd c_1$ and $\bd c_2$ were generated separately for the two datasets. $\bd X_1$ and $\bd X_2$ are shown in Figure \ref{fig:motivating_example}.

\paragraph{NIR of Corn:} The Cargill corn benchmark data was obtained from Eigenvector Research Inc.\footnote{http://www.eigenvector.com/data/Corn/ (accessed October 24, 2024)}. The dataset comprises NIR spectra from a set of 80 corn samples measured on 3 different spectrometers at 700 spectral channels in the wavelength range 1100-2498 \si{nm}. We used spectra from the m5 instrument along with the moisture contents in all experiments.  

\subsection{Evaluation}

\paragraph{Implementation Details:} $(\epsilon, \delta)$-differentially private PLS (edPLS) has been implemented with a scikit-learn compatible API using Python 3.10, featuring an EDPLS class with the usual ´fit()´ and ´predict()´ methods that integrate with pipelines to streamline preprocessing and modeling or the GridSearchCV class for hyperparameter optimization. The class is made available through the diPLSlib package, version 2.4.0., published at the Python Package Index (PyPI)\footnote{https://pypi.org/project/diPLSlib/} under a GPLv3 license. The usage pattern of the corresponding class is as follows:
\begin{verbatim}
>>> from diPLSlib import EDPLS
>>> import numpy as np
>>> x = np.random.rand(100, 10)
>>> y = np.random.rand(100,1)
>>> model = EDPLS(A=5, epsilon=0.1, delta=0.05)
>>> model.fit(x, y)
EDPLS(A=5, delta=0.05, epsilon=0.1)
>>> xtest = np.array([5, 7, 4, 3, 2, 1, 6, 8, 9, 10]).reshape(1, -1)
>>> yhat = model.predict(xtest)
\end{verbatim}
Code for reproducing the experiments has been made publicly available at \url{https://github.scch.at/circPlast-mr/-epsilon-delta--differentially-private-parital-least-squares-regression}\footnote{Note that due to the stochastic nature of the noise adding mechanism, results might differ slightly.}.

\paragraph{Simulated Data:} For the simulated data, we horizontally concatenated the spectral matrices $\bd X = [\bd X_1; \bd X_2]$ and analyte concentration vectors $\bd y = [\bd y_1, \bd y_2 $], and fitted models with 3 latent variables using $\epsilon = [100, 10, 1]$. In parallel, separate models (without noise addition) were fitted using $(\bd X_1, \bd y_1)$ and $(\bd X_2, \bd y_2)$.
Subsequently, we applied Gram-Schmidt orthogonalization as in Eq. \eqref{eq:w_perp} using the global and local weight matrices to estimate the unique source of variability present in the data contributed by the other data holder. Alternatively, we employed the same approach using the loadings matrix instead.

\paragraph{Corn Data:} We used spectra from the m5 instrument along with the moisture contents in all experiments. The dataset was randomly split into a training and test set ($70/30$). The training set was used to fit $(\epsilon, \delta)$-differentially private PLS models with different numbers of LVs using different values of $\epsilon$. We made extensive use of the ´GridSearchCV´ class from scikit-learn (v1.5.2) to perform (10-fold) cross-validation (CV) using the minimum root mean squared error (RMSECV) as the optimization criterion to choose the optimal number of latent variables (LVs) to retain in the models. The root mean squared error of prediction (RMSEP) on the test set was used as the utility function to assess the privacy-utility trade-off. We used the ´chemotools´ package (v.0.1.5) to perform spectral preprocessing using either the Savitzky-Golay (SG) filter (window size=5, order=2, derivative=1), airPLS ($\lambda=100$, iterations=15, order=1) or multiplicative scatter correction (MSC) using the mean spectrum as reference, and the ´PLSRegression´ class from scikit-learn to fit (baseline) PLS models for comparison.   

The failure probability $\delta$ was kept at $0.01(= 1\%)$  in all experiments.

\section{Results and Discussion}\label{results-and-discussion}

\subsection{Simulated Data}
\begin{figure}
    \centering
    \includegraphics[width=0.9\linewidth]{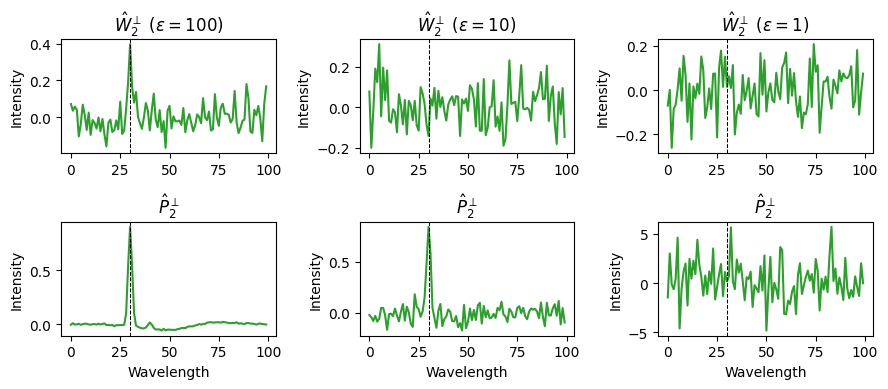}
\caption{Protection against privacy attacks. The recovery of the unique source of variability (encoded in LV 3) in the other data holder's data according to Eq. \eqref{eq:w_perp} is shown for decreasing privacy loss $\epsilon$ (at failure probability $\delta=0.01$) using the simulated dataset Figure \ref{fig:motivating_example}. The top and bottom plots show weights and loadings inferred by data holder 1, encoding unique sources of variability in data contributed by data holder 2. The dashed lines indicate the position of the unique peak in the other data holder's data (ground truth).}
    \label{fig:simulated_data}
\end{figure}

Figure \ref{fig:motivating_example} shows the recovery of the other data holder's unique source of variability encoded in LV 3 of an ordinary PLS model on the concatenated, simulated data set $(\bd X, \bd y)$. Data holder $\mathcal{D}_1$ (top) and $\mathcal{D}_2$ (bottom) can easily recover the unique peak (dashed line) in the other data holder's $X$-data by first obtaining the local weights from a PLS model on the local data $(\bd X_i, \bd y_i)$, followed by orthogonalization of the global weights with respect to the local weights according to Eq. \eqref{eq:w_perp}. The same "privacy attack" is shown for $(\epsilon, \delta)$-differentially private PLS models at decreasing privacy loss $\epsilon$ (from left to right) in Figure \ref{fig:simulated_data}. At $\epsilon = 100$, data holder one can reconstruct the unique peak present in data holder two's data holder data (leftmost plots). At $\epsilon = 10$, despite the recovery of the peak in the weights being hampered by the Gaussian noise, the corresponding attack using the loadings still succeeds. At $\epsilon=1$, reflective of strong $(\epsilon, \delta)$-differential privacy, the privacy attack on both weights and loadings is blunted. This example underpins that inferring (private) information present in the other data holder's data is blunted with strong privacy guarantees no matter which model components the CSP shares with the clients. Most notably, the DP framework is dataset independent, i.e., the corresponding privacy guarantees hold irrespective of the data or nature of the privacy attack. However, the natural question arises whether the utility of the model is high enough under (strong) privacy constraints or if the accuracy target can be met while maintaining some desired privacy level. To address these questions, in the next section, we investigated the trade-off between privacy and accuracy using the well-known NIR corn benchmark dataset. 

\subsection{NIR of Corn}
\begin{figure}[ht!]
    \centering
    \includegraphics[width=0.99\linewidth]{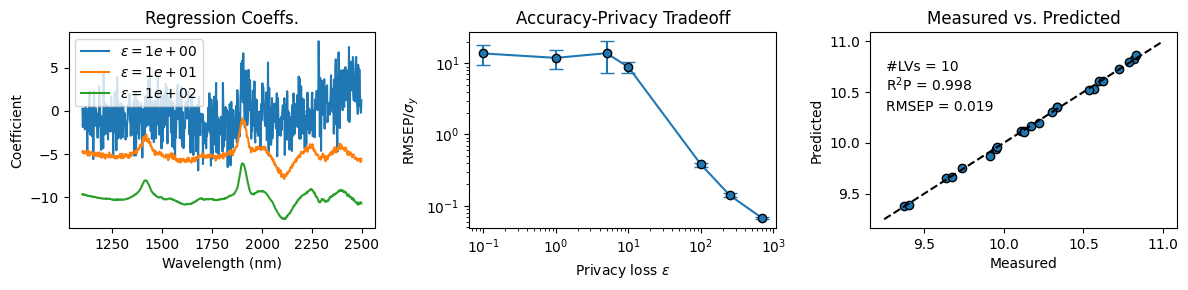}
    \caption{Privacy-accuracy trade-off.  Regression coefficients (left), the root mean squared error of prediction (RMSEP) divided by the standard deviation of $\bd y$ ($\sigma_y$) against privacy loss $\epsilon$ (middle) of edPLS models, and test set predictions of the baseline PLS model with 10 LVs (right) are shown. $R^2P$ indicates the $R^2$ score for the test set.}
    \label{fig:corn}
\end{figure}

Figure \ref{fig:corn} shows the accuracy-privacy trade-off of $(\epsilon,\delta)$-differential private PLS when decreasing the privacy loss parameter $\epsilon$ on the corn benchmark dataset for models with 10 latent variables. As expected, when $\epsilon$ is decreased, more noise is infused into the regression coefficients of the corresponding model (left plot). Concomitantly, the root mean squared error of prediction (RMSEP) on the test set increases, reminiscent of the accuracy-privacy trade-off in differential privacy (middle plot). The decrease in accuracy is particularly prominent below $\epsilon = 100$, indicating that the corresponding calibration does not admit strong privacy guarantees. Note that such strong guarantees require a high probability that the (PLS) algorithm yields identical models when trained on (any) neighboring datasets. This is the case when $\epsilon\leq 1$. 

\begin{figure}
    \centering
    \includegraphics[width=0.99\linewidth]{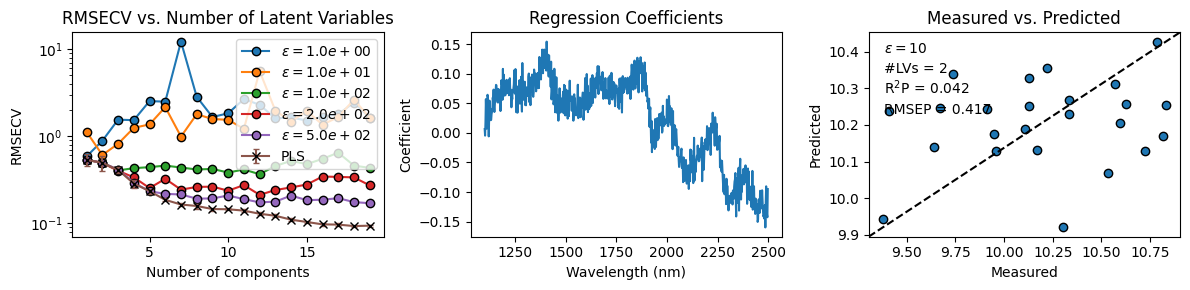}
    \caption{Model selection. The left plot shows the (10-fold) RMSECV vs. the number of LVs for edPLS models with different values of $\epsilon$ ($\delta=0.01$). Errorbars show the mean RMSECV for the corresponding baseline PLS models with the standard error of the mean across 100 experiments. Middle: Regression coefficients for an edPLS model with 2 LVs and $\epsilon=10$. Right: Test set predictions of the best model at $\epsilon=10$.}
    \label{fig:model_selection}
\end{figure}

 We next undertook a $10$-fold cross-validation to optimize the number of latent variables (LVs) to retain in the edPLS model. Figure \ref{fig:model_selection} (left) shows the root mean squared error of cross-validation (RMSECV) for different numbers of LVs and values of the privacy loss parameter $\epsilon$ (while keeping $\delta$ fixed at $0.01$). As expected, when $\epsilon$ is decreased, the overall error across different numbers of LVs increases. Furthermore, the smaller $\epsilon$, the smaller the optimal number of LVs, and overfitting becomes more pronounced as more LVs are retained, which is due to the accumulation of noise in the model. At $\epsilon = 10$, only two latent variables are optimal, whereas, at $\epsilon = 1$, the RMSECV increases beyond 1 LV, confirming that no meaningful model with strong privacy guarantees can be derived. The middle plot of Figure \ref{fig:model_selection} shows the regression coefficients for an edPLS model with $\epsilon=10$ and two LVs, and the rightmost plot shows the corresponding test set predictions indicative of a poor calibration model for the determination of moisture in corn.

\begin{figure}
    \centering
    \includegraphics[width=0.99\linewidth]{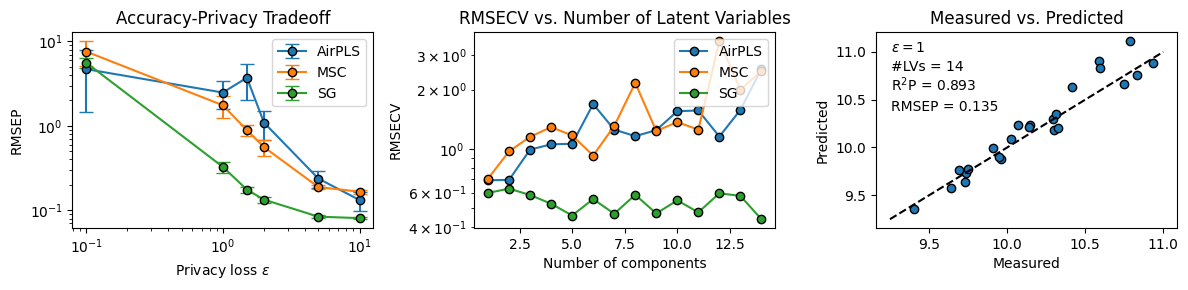}
    \caption{Privacy-preserving spectral pre-processing. Left: Root mean squared error of prediction (RMSEP) on the test set against privacy loss $\epsilon$ based on pre-processed spectra. AirPLS: Adaptive iteratively reweighed penalized least squares baseline correction, MSC: Multiplicative Scatter Correction, SG: Savitzky-Golay (1st derivative) filter. The number of LVs was kept at 8. Middle: The root mean squared error of cross-validation (RMSECV) against the number of LVs for $\epsilon=1$. Right: Test test predictions of the optimal model at $\epsilon=1$ based on SG-preprocessed spectra.}
    \label{fig:preprocessing}
\end{figure}

(NIR) spectra are often distorted due to scattering. The corresponding spectral artifacts (e.g., baselines) usually contribute to variance in $X$ but not in $Y$. Removing such non-predictive sources of variability beforehand usually increases the similarity between individual spectra and should thus enable stronger privacy guarantees (at acceptable deterioration of accuracy). To test this hypothesis, we employed three widely used pre-processing approaches in spectroscopy to pretreat the corn spectra before applying edPLS: Baseline removal using adaptive iteratively reweighed penalized least squares (airPLS) baseline correction, multiplicative scatter correction (MSC) and first derivative Savitzky-Golay (SG) filtering. Figure \ref{fig:preprocessing} (left) shows the RMSEP against the privacy-loss for the corresponding pre-processing methods. In general, SG-based pre-processing exhibits the best accuracy-privacy trade-off among the three methods, whereas airPLS and MSC yield similar accuracy on the test set across different values of $\epsilon$. We next optimized the number of LVs for edPLS models with $\epsilon=1$ (i.e., strong privacy guarantees) by means of cross-validation. The middle plot in Figure \ref{fig:preprocessing} shows the RMSECV against the number of LVs for the three pre-processing methods. Notably, the RMSECV only decreased upon the inclusion of more than 1 LV for models based on SG pre-processed spectra, indicating that MSC and airPLS don't admit strong privacy guarantees for the corn data set. The optimal edPLS model using SG pre-processed spectra at $\epsilon=1$, on the other hand, still exhibits reasonable accuracy on the test set while providing strong privacy guarantees (Figure \ref{fig:preprocessing}, right).             

\section{Conclusions}\label{conclusions}
In the present study, we have designed Gaussian noise adding mechanisms, according to \cite{dwork2014algorithmic} and \cite{balle18a} for the different steps of the PLS1 algorithm to ensure $(\epsilon, \delta)$-differential privacy of the final model (components). Strong differential privacy provably prevents the reconstruction of the training data from statistical models, rendering any sort of privacy attack ineffective. We demonstrated, based on simulated data, that under strong privacy guarantees ($\epsilon \leq 1$), privacy attacks aimed at recovering unique sources of variability fail. We then tested edPLS on the corn benchmark and found that achieving strong privacy and acceptable utility based on raw spectra is difficult. Proper selection of the pre-processing method, however, allowed optimization of the privacy-accuracy trade-off significantly, achieving either strong privacy or decent accuracy. With edPLS, we provide a practical tool to enable the creation of privacy-preserving multivariate calibrations and to analyze their privacy-utility trade-offs. Yet, we would like to stress that in federated multivariate calibration settings, edPLS requires sharing potentially sensitive (raw) data with a computation service provider, which might be problematic due to privacy concerns. Privacy enhancing techniques like, e.g., the random masking approach proposed by Nguyen Duy \emph{et al.} in \cite{nguyen2023fedmspc}, could be integrated with edPLS, thus safeguarding the privacy across the entire process - from data transmission to the release of the differential private model. We leave the required developments for future research.  

\section{Acknowledgement}\label{acknowledgement}

This contribution has been funded by BMK, BMAW, and the State of Upper
Austria in the frame of the SCCH competence center INTEGRATE (FFG grant
no. 892418) part of the FFG COMET Competence Centers for Excellent
Technologies Program, the COMET Center CHASE.

\newpage
\printbibliography[title={References}]

\section*{Appendix}

For the  derivation of the upper bound on the sensitivity of $\bd w(\bd X, \bd y)$ we first note that the $\ell_2$-norm difference between neighboring datasets $D=(\bd X, \bd y)$ and $D'=(\bd X', \bd y')$, given the definition in section \ref{sec:notation}, is
\begin{align}
\begin{aligned}
\|\bd X^T\bd y - \bd X'\bd y'\|_2 &= \|\bd x_n y_n - \sum_{j=1}^{n-1}(\bd x_j y_j - \bd x'_j y'_j)\|_2\\
& = \|\bd x_n y_n\|_2,
\end{aligned}
\end{align}
i.e., the worst case $\ell_2$-norm difference between neighboring datasets solely depends on the difference in the $n$-th entry on which they differ. Exploiting the fact that $\|\bd{a} b\|_2=\|\bd{a}\|_2|b|$, we have

\begin{align}
\label{eq:w_sensitivity_app}
\begin{aligned}
\Delta w &= \sup_{D, D'} \|\bd w(\bd X, \bd y) - \bd w(\bd X', y')\|_2 \\
         &= \sup_{(\mathbf x, y)} \|\bd x  y\|_2 \\
         &= \sup_{(\mathbf x, y)} |y| \|\bd x\|_2.
\end{aligned}
\end{align}
The scores vector $\bd t$ has $n$ and $n-1$ entries for neighboring datasets $D$ and $D'$, respectively. Thus the sensitivity $\Delta t$ is equal to the (worst-case) $\ell_2$-norm of the $n$-th entry, i.e., 

\begin{align}
\label{eq:t_sensitivity_app}
\begin{aligned}
\Delta t &= \sup_{\mathbf x} \|\bd x^T\tilde{\mathbf{w}}\|_2\\
         &\leq \sup_{\mathbf x} \|\bd x\|,
\end{aligned}
\end{align}
with $\tilde{\mathbf{w}} = \mathbf{w}/\|\mathbf{w}\|_2$. The upper bound follows from the Cauchy-Schwarz inequality ($\|\bd a^T\bd b\|\leq\|\bd a\|\|\bd b\|$) and $\|\tilde{\mathbf{w}}\|_2=1$. Similarly, the sensitivity of $\bd p(\bd X, \tilde{\bd t})$, with $\tilde{\bd t} = \bd t/\|\bd t\|_2$ depends solely on the norm of the $n$-th entry, i.e., 

\begin{align}
\label{eq:p_sensitivity_app}
\begin{aligned}
\Delta p &= \sup_{D, D'} \|\bd p(\bd X, \tilde{\bd t}) - \bd p(\bd X', \tilde{\bd t}')\|_2\\
         &= \sup_{D, D'} \|\tilde{t}_n \bd x_n - \sum_{j=1}^{n-1}(\tilde{t}_j\bd x_j- \tilde{t}'_j\bd x'_j)\|_2\\
         &= \sup_{D, D'}\|\tilde{t}\bd x\|_2\\
         &\leq \sup_{\mathbf x}|\|\bd x\|_2.
\end{aligned}
\end{align}
The inequality follows from $\|\tilde{t}\bd x\| = |\tilde{t}|\|\bd x\|$ and $|\tilde{t}|\leq 1$ due to $\|\tilde{\bd t}\|_2 = 1$. Following the same arguments as above the sensitivity of the $Y$- loadings
\begin{align}
\label{eq:c_sensitivity_app}
\begin{aligned}
\Delta c &= \sup_{D, D'} \|c(\bd y, \tilde{\bd t}) - c(\bd y', \tilde{\bd t}')\|_2\\
         &= \sup_{D, D'} |\tilde{t} y|\\
         &\leq \sup_{y} |y|.
\end{aligned}
\end{align}

\end{document}